\documentclass[conference]{IEEEtran}
\usepackage{times}
\usepackage{color}
\usepackage{graphicx}
\usepackage[dvipsnames]{xcolor}
\usepackage[numbers]{natbib} % numbers option provides compact numerical references in the text. 
\usepackage{multicol}
\usepackage[bookmarks=true]{hyperref}
\usepackage{multirow}
\usepackage{makecell}
\usepackage{wrapfig}
\usepackage{booktabs}
\usepackage{amsmath}

\definecolor{myred}{HTML}{EB5270} % Hex color code
\definecolor{mygreen}{HTML}{00D0A1} % Hex color code
\definecolor{myblue}{HTML}{6491EA} % Hex color code

\begin{document}

% \title{Potential Surprise:\\ A Lens for Unveiling Interactive Driving Scenarios}
% \title{Potential Surprise as a Lens for Unveiling Interactive Driving Scenarios}
% \title{Investigating Surprise Potential as a Measure of Interactivity in Driving Scenarios}
% \title{Surprise Potential as a Measure of Interactivity in Driving Scenarios: A Case Study}
\title{Surprise Potential as a Measure of \\ Interactivity in Driving Scenarios}

% You will get a Paper-ID when submitting a pdf file to the conference system
% \author{Author Names Omitted for Anonymous Review. Paper-382}

\author{\authorblockN{Wenhao Ding$^{1}$\ \ Sushant Veer$^{1}$\ \ Karen Leung$^{1,2}$\ \ Yulong Cao$^{1}$\ \ Marco Pavone$^{1,3}$}
\authorblockA{$^{1}$NVIDIA Research\ \ $^{2}$University of Washington\ \ $^{3}$Stanford University \\
% Corresponding Email: {wenhaod@nvidia.com}
}
}

% avoiding spaces at the end of the author lines is not a problem with
% conference papers because we don't use \thanks or \IEEEmembership

%Homer Simpson\authorrefmark{2},
%Montgomery Scott\authorrefmark{3} and
%Eldon Tyrell\authorrefmark{4}}
%Telephone: (800) 555--1212, Fax: (888) 555--1212}
%\authorblockA{\authorrefmark{4}Tyrell Inc., 123 Replicant Street, Los Angeles, California 90210--4321}}

\maketitle

\begin{abstract}
Validating the safety and performance of an autonomous vehicle (AV) requires benchmarking on real-world driving logs. However, typical driving logs contain mostly uneventful scenarios with minimal interactions between road users. Identifying interactive scenarios in real-world driving logs enables the curation of datasets that amplify critical signals and provide a more accurate assessment of an AV's performance. In this paper, we present a novel metric that identifies interactive scenarios by measuring an AV's \textit{surprise potential} on others. First, we identify three dimensions of the design space to describe a family of surprise potential measures. Second, we exhaustively evaluate and compare different instantiations of the surprise potential measure within this design space on the nuScenes dataset. To determine how well a surprise potential measure correctly identifies an interactive scenario, we use a reward model learned from human preferences to assess alignment with human intuition. Our proposed surprise potential, arising from this exhaustive comparative study, achieves a correlation of more than 0.82 with the human-aligned reward function, outperforming existing approaches. Lastly, we validate motion planners on curated interactive scenarios to demonstrate downstream applications.
\end{abstract}

\IEEEpeerreviewmaketitle

\section{Introduction}\label{sec:intro}

\begin{figure}[t]
\centering
\includegraphics[width=0.99\linewidth]{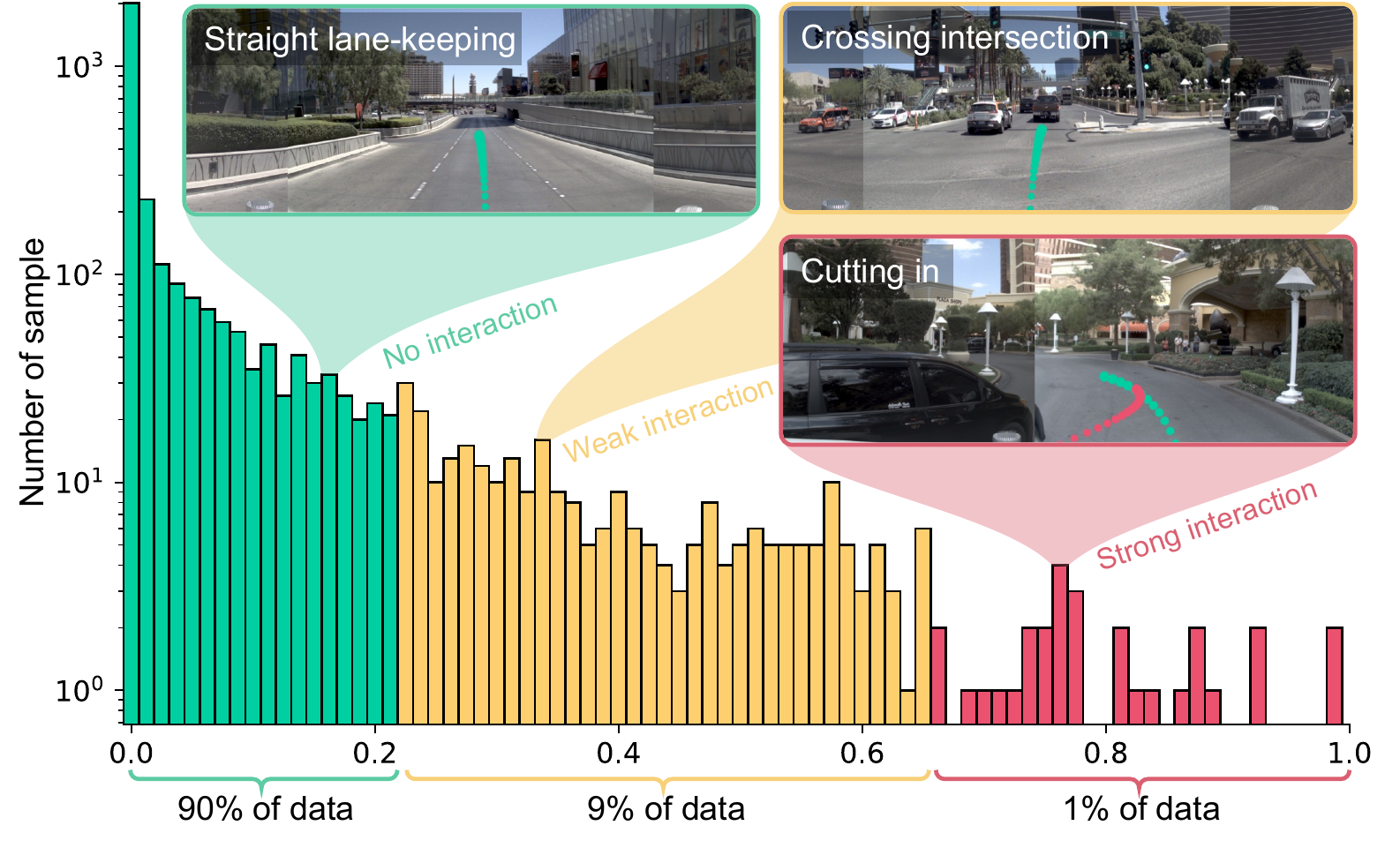}
\vspace{-7mm}
\caption{
The distribution of the interactive score obtained by surprise potential on the nuScenes dataset~\cite{nuscenes}. We notice that most scenarios in the dataset are not considered interactive.}
\vspace{-6mm}
\label{fig:teaser}
\end{figure}

% the motivation of discovering interactive scenarios
Autonomous vehicles (AV) are highly dependent on real-world driving logs for software development, verification, and validation. However, the majority of recorded driving data consists of uneventful driving with limited interactions~\cite{dauner2023parting}, offering insufficiently challenging benchmarks. 
For example, the nuScenes dataset~\cite{nuscenes} exhibits a highly imbalanced distribution of interactive scenarios, as illustrated in Figure~\ref{fig:teaser}, where less than 10\% of the scenarios involve significant interactivity.
This imbalance can distort performance metrics, thus obscuring a clear assessment of the AV stack’s safety and capabilities~\cite{ding2023survey}. To effectively test the AV stack and improve its ability to handle critical driving situations, it is essential to curate and evaluate interactive driving scenarios.

% the explanation of using surprise as a metric for interaction
Indeed, the challenge of identifying highly interactive scenarios has attracted significant interest, leading to various approaches. Broadly, these methods can be categorized into rules-based~\cite{sadat2021diverse}, supervised learning~\cite{bronstein2023embedding}, and unsupervised learning~\cite{dinparastdjadid2023measuring} approaches. 
Rule-based methods have limited expressivity to describe complex interactions, while supervised learning methods require substantial human effort. Therefore, in this paper, we lean towards unsupervised learning methods.

In particular, we explore the curation of interactive scenarios by quantifying their \emph{surprise potential}\footnote{Prior literature refers to this as surprise while omitting the term ``potential." However, the methods discussed in this paper and prior works, such as \cite{dinparastdjadid2023measuring}, estimate the \emph{potential} for surprise rather than the actual surprise of a scenario.}, which is generally aligned with human experience~\cite{dinparastdjadid2023measuring, tolstaya2021identifying, engstrom2024modeling}. 
In the context of driving, when an agent is surprised by the behavior of another, the surprise manifests itself in the form of a sudden and significant deviation from the expected \emph{nominal behavior}, for example, sudden braking or swerving. Motivated by this observable outcome of being surprised, a widely adopted approach -- also employed in this paper -- for estimating surprise potential involves counterfactual reasoning. 
The trajectory of an agent in the scene is perturbed and the distribution shift in the behavior of other agents is measured \cite{dinparastdjadid2023measuring, tolstaya2021identifying, engstrom2024modeling} -- the larger the distribution shift, the more surprising a scenario can be. The surprise potential of a scenario can be likened to potential energy, which, when unleashed, can provoke surprising reactions from other traffic agents, potentially compromising safety.

% main content and method of this paper
In this paper, we present a detailed comparative study on the class of surprise potential estimation methods that leverage counterfactual reasoning~\cite{lewis2013counterfactuals}. We present insights into the design choices underlying these methods from three perspectives: (i) the method for counterfactual generation; (ii) the architecture of the future prediction model; and (iii) the distance metric for evaluating distribution shift. This comprehensive investigation enables us to develop new instances that modify history trajectories to predict counterfactual future scenarios, and employ the Wasserstein distance to measure the distribution shift between these scenarios.

% we also have some novelty in the evaluation part
The absence of ground-truth interactivity labels presents a significant challenge in evaluating the performance of different surprise potential methods. Given the difficulty of assigning absolute interactivity scores, we developed a tool to collect human preference labels and employed ranking-based correlation metrics for evaluation. Beyond human evaluation, we also demonstrate that interactive scenarios identified using these metrics can benefit downstream planning models, highlighting the practical value of surprise metrics in autonomous driving.

% highlight the contribution
\textbf{Statement of Contributions.} We make four main contributions in this paper:
\begin{enumerate}
    \item We propose a family of surprise metrics by decomposing the design space into three dimensions and identifying multiple instantiations to explore, which differ in the counterfactual perturbation of the scenario, the architecture of the future prediction model, and the metrics for measuring distribution shift.
    \item We perform an exhaustive evaluation to characterize the various aspects of the design space and assess the performance of these metrics through human alignment.
    \item Based on our comparative findings, we present a novel surprise metric, a hitherto unexplored combination of design dimensions. The new approach achieves a correlation that exceeds 0.82 with human labels.
    \item We show that using an interactive dataset curated by our metric significantly benefits downstream applications, such as planning. The insights of this paper are transferrable more broadly to any robotics application that involves multi-agent interactions.
\end{enumerate}

\begin{figure*}[t]
\centering
\includegraphics[width=0.99\linewidth]{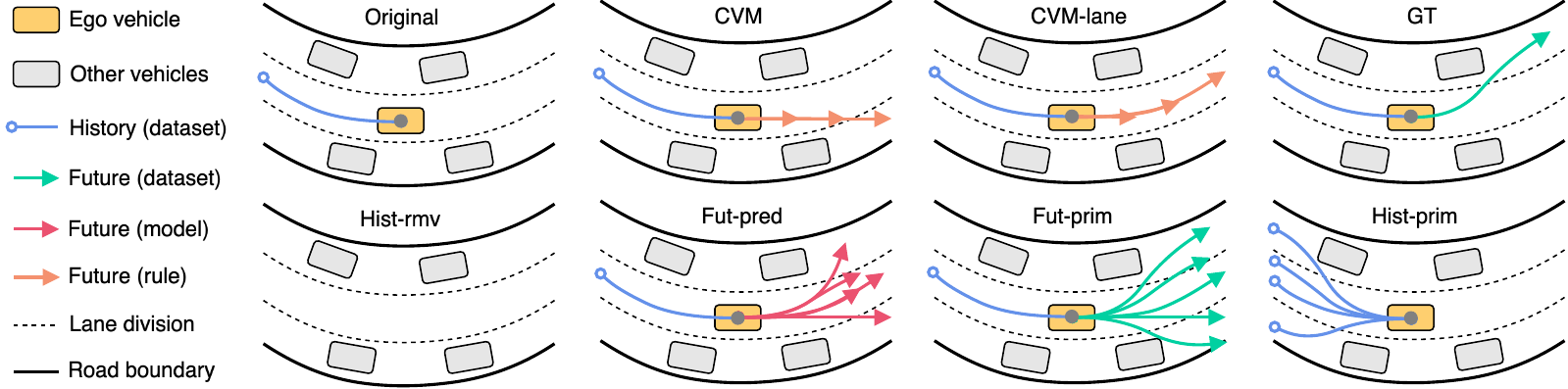}
\vspace{-1mm}
\caption{The illustration of counterfactual generation methods considered in this paper.}
\vspace{-4mm}
\label{fig:metric}
\end{figure*}

\section{Related Work}

We discuss various approaches to determining the ``interestingness'' of a scenario, a concept that can vary depending on the use case. ``Interestingness'' can refer to the level of interactivity between agents, the safety-critical nature of the scenario, or the predictability of agent behavior. Identifying such ``interesting'' scenarios enables the curation of more challenging and informative datasets. We also provide a concise overview of trajectory prediction models, which are key components in analyzing the interestingness of a scenario.

\subsection{Surprisingness of an outcome}
The concept of surprise in the context of AV is introduced in \cite{dinparastdjadid2023measuring,engstrom2024modeling}. Although various options have been proposed to quantify surprise, it is generally understood as a comparison between the posterior distribution or outcome and the prior distribution. 
When the posterior or outcome significantly deviates from what the prior distribution describes, the scenario is considered surprising, as it was not adequately anticipated by the prior. A key insight from these studies is that there is no universal formula for measuring surprise. Instead, multiple mathematical approaches and interpretations can be employed. This paper builds upon the concept of surprise, aiming to investigate and evaluate a family of potential surprise measures and assess their effectiveness based on human preferences.

\subsection{Measuring interactivity using counterfactual analysis}
A scene can be considered interesting if the behavior of one agent directly influences the behavior of another, indicating that the agents are \textit{interacting}. However, close proximity does not necessarily imply interactivity. As such, a commonly used approach to measure interactivity involves counterfactual analysis to determine how differently an agent behaves if another agent acts differently. Using powerful trajectory prediction models, \cite{tolstaya2021identifying} measures interactivity by calculating the Kullback–Leibler (KL) divergence between a target agent's (i) future distribution conditioned on a query agent's future trajectory and (ii) its future marginal distribution. The KL divergence is not symmetric and other divergence metrics could be more applicable. Rather than conditioning on a query agent's future, \cite{hsu2023interpretable, schaefer2021leveraging} instead removes the query agent from the scene to determine whether the presence of one agent affects the behavior of another. Task relevance can also be incorporated by considering the distribution of rewards rather than trajectories, as discussed in \cite{stoler2024safeshift}. However, this approach requires knowledge of each agent's desired task, which may be challenging to obtain or infer.

\subsection{Dynamics-based identification of interesting scenarios}
There are various methods that reason about the dynamics of agents when determining whether a scenario is considered interesting. For instance, \cite{makansi2021exposing,feng2024unitraj} evaluate scenario interestingness, particularly in terms of difficulty, by analyzing the accuracy of a linear prediction model, such as a Kalman filter. If the Kalman prediction is poor, a non-linear prediction becomes necessary, implying the scenario is ``interesting.'' Alternatively, some methods employ basic extrapolation (e.g., constant velocity, braking, optimal control) to determine whether agents are likely to collide in the near future~\cite{sadat2021diverse,jiang2024interhub,topan2022interaction,westhofen2023criticality,junietz2018criticality}.
While these dynamics-based approaches are relatively straightforward and do not rely on large datasets, they involve simplifying assumptions that fail to capture non-trivial behaviors influenced by environmental factors such as road geometry and traffic rules. In contrast, rather than explicitly modeling dynamics, \cite{bronstein2023embedding} uses supervised learning with driving logs to estimate collision potential. This method predicts the likelihood of a collision in the near future based on previous trajectory segments and environmental context but requires access to extensive and diverse datasets.

\subsection{Motion Prediction Models}
AV trajectory predictors can be classified based on how agent trajectories are represented (agent-, scene-, or query-centric) and the network architecture employed (Transformer, diffusion model, or autoregression). In agent-centric representations (e.g.~\cite{zhao2021tnt,gu2021densetnt,shi2022motion}), the model generates marginal predictions for a single target agent, treating other agents as part of the environment. This approach facilitates linear runtime by making separate inferences for each agent, which complicates the derivation of joint predictions. Scene-centric representation, on the other hand, predicts the motion of all agents jointly using a single representation relative to a central reference agent, requiring only one inference step. However, it tends to sacrifice prediction accuracy for agents located farther from the ego vehicle due to inherent spatial biases. Query-centric models~\cite{zhou2023query,shi2024mtr++} combine aspects of agent- and scene-centric representations. Although requiring more memory, it strikes a balance by being more efficient than agent-centric models and more accurate than scene-centric ones. Since our surprise metric relies on the joint distribution of future trajectory, this work focuses on scene- and query-centric models.

From a network architecture standpoint, trajectory prediction models differ in how trajectory predictions are produced. 
Transformer-based models (e.g.,~\cite{ngiam2021scene,girgis2021latent}) typically encode input features and then decode the future trajectory in a single forward step. Diffusion-based models~\cite{ho2020denoising} are gaining popularity for their ability to model complex distributions, leveraging a guidance function to influence the denoising process~\cite{zhong2023guided,zhong2023language,huang2024versatile,jiang2023motiondiffuser}. Inspired by advances in natural language community, autoregressive models~\cite{zhang2024closed,philion2024trajeglish,wu2024smart,hu2025solving,zhao2024kigras,seff2023motionlm} tokenize trajectory data, effectively converting it into a discrete representation. However, this discretization complicates the measurement of distributional shifts. As such, this work focuses on diffusions and transformers, specifically those with single-step forward structures.

\section{Problem Formulation}

Our primary objective is to identify scenarios that humans are likely to perceive as interactive. To achieve this, we propose using surprise potential as an indicator of the interactivity of driving scenarios. In this section, we first provide a concrete definition of surprise potential and decompose it into three key dimensions, which are analyzed in detail. Finally, we outline the evaluation protocol employed in this study to assess the proposed surprise potentials.

\subsection{Definition of Surprise Potential}
Traffic agents exhibit surprise through significant deviations from their ``nominal" motion, i.e., their intended motion, in response to changes in their environment. For example, a driver may abruptly brake, thus deviating from their desired motion of maintaining speed, due to an adjacent driver swerving in front of them. However, scenarios where agents are genuinely surprised are relatively rare. Rather than estimating the actual surprise of a scenario, we focus on the \textit{potential} for surprise. We measure the surprise potential of a scenario by estimating how large the deviations would have been for an agent in the presence of a counterfactual (e.g., a vehicle suddenly braking). In the counterfactual scenario, minor changes in the trajectories of other agents from the nominal scenario suggest a low surprise potential, while large changes imply a high surprise potential. 

Next, we introduce the surprise potential in a mathematical formulation. Let $\xi \in \Xi$ denote a traffic scenario encompassing the historical and future trajectories of all agents, their bounding boxes, agent types, map elements, and any other factors relevant to the decision-making of traffic agents. Mathematically, we define the Surprise Potential (SP) as a function $S:\Xi \to [0,\infty)$ that outputs a non-negative scalar, where the magnitude reflects the degree of surprise associated with the scenario. We study a general class of SPs that measure the distribution shift in agent trajectories under counterfactual interventions. 
Let $\mathcal{F}:\Xi \to \mathcal{X}$ be a trajectory prediction model that given a scene $\xi \in \Xi$ outputs a distribution $x\in\mathcal{X}$ for the trajectories of all agents in a scene. Let $\mathcal{G}:\Xi \to \Xi$ be a counterfactual scenario generator that ``edits" an input scenario to synthesize an altered one. Finally, let $\mathcal{D}:\mathcal{X}\times\mathcal{X}\to[0,\infty)$ measure distribution shifts (e.g., KL-divergence) between two trajectory distributions. Then, we express the SP as:
\begin{equation}
    S(\xi) := \mathcal{D}(\mathcal{F}(\xi), \mathcal{F}\circ\mathcal{G}(\xi)).
\label{equ:surprise}
\end{equation}

\subsection{Design Space Decomposition}\label{subsec:design-space-dim}

According to the definition in~\eqref{equ:surprise}, an instance of the surprise potential consists of three main components:
\begin{itemize}
    \item \textbf{Counterfactual generation function $\mathcal{G}$}. This function can be a rule-based, search-based, or even learning-based method, leading to either minor perturbations to the original trajectory or a totally different trajectory.
    \item \textbf{Future prediction model $\mathcal{F}$}. Leveraging advancements in motion prediction, well-established models can be utilized to forecast future scenarios effectively. These models provide the necessary trajectory distributions for evaluating SP.
    \item \textbf{Distribution shift measurement $\mathcal{D}$}. This metric quantifies the differences between potential future trajectories, which directly impacts the scale and distribution of the surprise metric by evaluating how the scenarios deviate under counterfactual conditions. 
\end{itemize}
We investigate design choices for each of these components and evaluate the performance of different configurations.

\subsection{Metrics for Evaluation}

To compare different designs of the surprise potential, a suitable metric is necessary. Unfortunately, ground-truth labels indicating the interactivity of scenarios are typically unavailable, as obtaining such labels is both costly and challenging for human annotators, who would struggle to assign absolute interactivity scores to driving scenarios. To address this issue, we propose an alternative approach: collecting pairwise preferences between scenarios, which can then be used to derive a ranking of all scenarios within the dataset. Despite this, ranking the \textit{entire} dataset directly is computationally intractable, as it calls for annotating $O(N^2)$ pairs for a dataset containing $N$ scenarios. To overcome this limitation, we train a reward model using a few annotated pairs and use it to predict rankings for the entire dataset.

\section{The Instantiations of Surprise Potential}\label{sec:surprise-instances}

This section discusses the various instantiations of surprise potential that emerge from different configurations of the design space, as outlined in Section~\ref{subsec:design-space-dim}. The proposed design space encompasses certain existing approaches while also introducing novel measures of surprise potential.

\subsection{Counterfactual Scenario Generation Function}

We begin with straightforward manipulations of the recorded trajectories. \textbf{Fut-none} represents the standard motion prediction setting, where the model uses only the historical trajectories of all agents as input. \textbf{Fut-gt} adopts a conditional prediction approach, incorporating the ground-truth future trajectory of the target agent as a condition for predicting the future trajectories of other agents~\cite{huang2023conditional}. \textbf{Hist-rmv} modifies the past by removing the target agent's history, effectively simulating a scenario in which the target agent does not exist~\cite{hsu2023interpretable,schaefer2021leveraging}.

Beyond manipulating the recorded data to generate counterfactual scenarios, we explore more sophisticated counterfactuals that utilize the dynamics model and lane information. \textbf{Fut-cvm} conditions on the future trajectory of the target agent generated by a constant velocity dynamics model. \textbf{Fut-cvm-l} extends this approach by employing a lane-following dynamics model to generate the future trajectory of the target agent based on the closest lane, which is then used as a condition.

We can also use learning-based future predictions for agents as counterfactuals by leveraging a multi-modal marginal motion prediction model. This approach, inspired by prior work~\cite{tolstaya2021identifying}, is referred to as \textbf{Fut-pred}. The predicted future trajectories are used as conditions for the predictor $\mathcal{F}$ in computing the surprise potential~\eqref{equ:surprise}.

Inspired by motion prediction models that represent trajectories with tokens~\cite{wu2024smart}, we introduce two perturbations utilizing motion primitives from the dataset as counterfactuals. Motion primitives are short trajectory segments of a single agent, spanning several seconds, that encapsulate behavior-level information such as left turns or lane changes. Specifically, we define two counterfactuals: \textbf{Hist-prim}, which replaces the target agent's ground-truth history with motion primitives, and \textbf{Fut-prim}, which substitutes the ground-truth future with motion primitives.
To ensure the feasibility of these primitives in given scenarios, we only select those that do not result in collisions with other agents and remain within drivable areas. All counterfactual generation functions discussed here are illustrated in Figure~\ref{fig:metric}.

\subsection{Future Prediction Model}

For the future prediction model, we consider two essential factors. First, the model must support both unconditional and conditional prediction, as both scenarios are required for our counterfactuals. Second, the model should produce a multi-modal distribution to effectively capture the inherent uncertainty in the behavior of traffic agents.

Inspired by recent advances in motion predictors~\cite{zhou2023query,jiang2023motiondiffuser}, we investigate two key design choices: scene representation and model architecture. For scene representation, we evaluate two approaches: scene-centric (\textit{SC}) and query-centric (\textit{QC}). Regarding the model architecture, we consider a feedforward prediction architecture (\textit{FFP}) and a diffusion-based architecture (\textit{Diff}). The \textit{FFP} model directly predicts the $(x, y)$ locations, while the \textit{Diff} model predicts acceleration and yaw rate. To incorporate multimodality, \textit{FFP} uses a Gaussian Mixture Model (GMM) as its prediction head, whereas \textit{Diff} employs a sample-based distribution. In total, we evaluate four models: \textbf{FFP-SC}, \textbf{FFP-QC}, \textbf{Diff-SC}, and \textbf{Diff-QC}.

\subsection{Distribution Shift Measurement}

For the last dimension of the surprise metric, we consider various metrics to measure the distribution shift between two GMMs $\sum_{k=1}^K \pi_1^k \mathcal{N}(\mu_1^k, \Sigma_1^k)$ and $\sum_{k=1}^K \pi_2^k \mathcal{N}(\mu_2^k, \Sigma_2^k)$ from the prediction model: $L^2$-norm (\textbf{L2}) in the Euclidean space between the means of top-$K$ modes, KL divergence (\textbf{KLD}), and Wasserstein distance (\textbf{W2}).

\textbf{L2 norm.} First, we consider a simple yet efficient $L^2$-norm in the Euclidean space. We sort the modes according to their probabilities in the GMM and then calculate the L2 distance between the mean values of the matched modes:
\begin{equation}
    \mathcal{D}_{L2} = \sum_{k=1}^{K} \|\mu_1^k - \mu_2^k \| ^{1/2}
\end{equation}

\textbf{KLD.} We then consider the KL divergence (KLD), which is widely used to measure the discrepancy between two distributions~\cite{tolstaya2021identifying}. Since there is no closed-form equation for the KLD of two GMMs, we use sample-based methods for estimation:
\begin{equation}
    \mathcal{D}_{KLD} = \frac{1}{N} \sum_{i=1}^N \log \frac{\sum_{k=1}^K \pi_1^k \mathcal{N}(x_i \mid \mu_1^k, \Sigma_1^k)}{\sum_{k=1}^K \pi_2^k \mathcal{N}(x_i \mid \mu_2^k, \Sigma_2^k)}.
\end{equation}

\textbf{W2 distance.} We notice that L2 ignores the probability of mode $\pi$ and KLD only measures relative information loss without accounting for the geometric structure of the underlying space. To measure the discrepancy between multi-modal prediction, we need a better metric that leverages the similarity between different modes. Therefore, we propose to use optimal transport~\cite{villani2009optimal} as a distribution shift measure. Specifically, we use the $2^{nd}$-order Wasserstein distance between GMMs, which involves two computation stages.
The first stage computes the cost matrix $M$ between every pair of Gaussian distribution modes, which has the following closed-form solution. Thus, we can get all elements $M_{ij}$ in the matrix:
\begin{equation}
M_{i,j} = \|\mu_i - \mu_j\|_2^2 + \mathrm{Tr}\left[\Sigma_i + \Sigma_j - 2 \left(\Sigma_i^{1/2} \Sigma_j \Sigma_i^{1/2}\right)^{1/2}\right].
\end{equation}
The second stage solves the linear programming formulation of the optimal transport problem, which minimizes the total cost of transporting the mass from one distribution to another. Specifically, the optimal transport problem is
\begin{equation}
    \begin{aligned}
        \mathcal{D}_{W2} = & \min_{\gamma \in \Gamma(\pi_1, \pi_2)} \sum_{i=1}^K \sum_{j=1}^K M_{i,j} \gamma_{i,j}, \\
        & \text{subject to} \sum_{j=1}^K \gamma_{i,j} = \pi_1^i, \sum_{i=1}^K \gamma_{i,j} = \pi_2^j, \gamma_{i,j} \geq 0,
    \end{aligned}
\end{equation}
where $M_{i,j}$ is the elements of the cost matrix and $\gamma_{i,j}$ is the transport plan, which represents the amount of mass transported from the $i$-th mode in $\pi_1$ to the $j$-th mode in $\pi_2$, and $\Gamma(\pi_1, \pi_2)$ is the space of the transport plans.

\begin{figure}[t]
\centering
\includegraphics[width=0.99\linewidth]{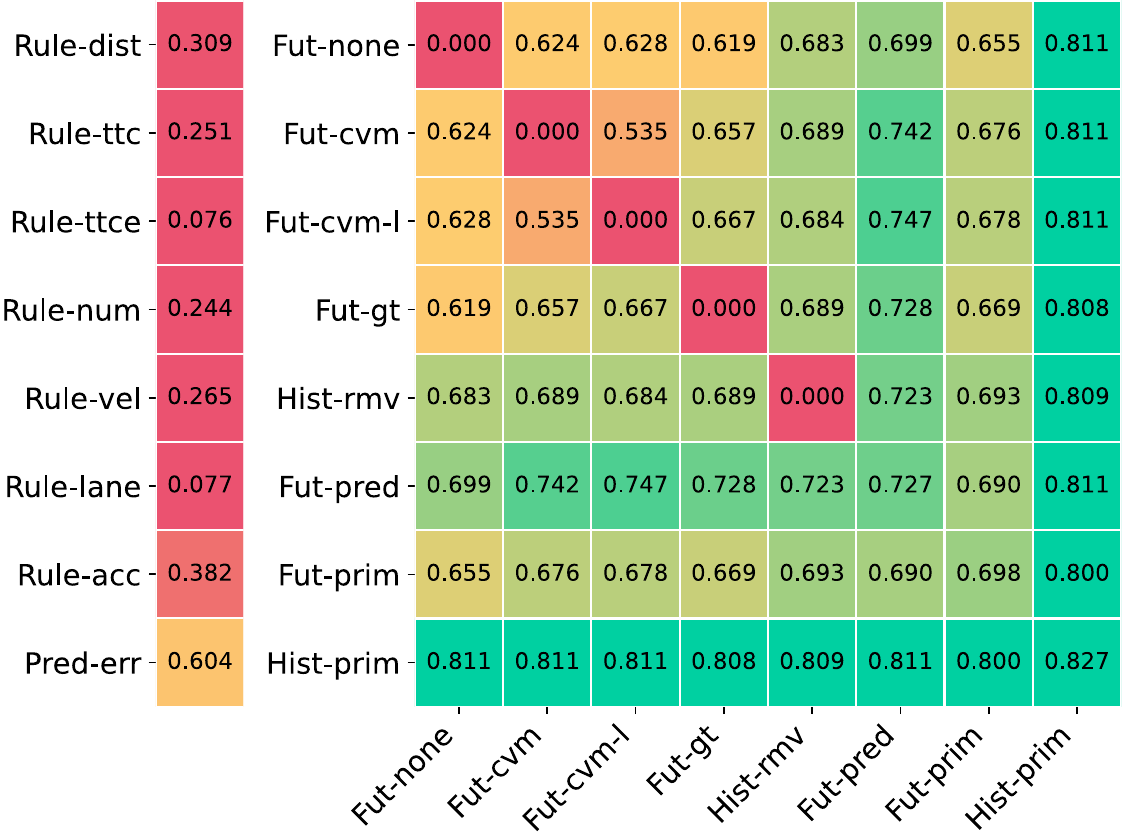}
\vspace{-7mm}
\caption{The Spearman rank correlation of various metrics for evaluating interactivity, where higher values represent better agreement.}
\vspace{-5mm}
\label{fig:matrix_corr}
\end{figure}

\section{Experimental set up and analysis}\label{sec:experiment}

In this section, we perform experiments to investigate the design space of surprise potentials outlined in Section~\ref{sec:surprise-instances}.

\subsection{Experimental Setup and Baselines}

In all of our experiments, we use the nuScenes dataset~\cite{nuscenes}. Trajectories are divided into segments consisting of 5 seconds of historical data and 4 seconds of future data. The variations of trajectory predictors employed in this study are based on a spatial-temporal transformer architecture~\cite{waswani2017attention} and are trained from scratch. The primary model used for both quantitative and qualitative evaluation is referred to as \textit{FFP-QC-W2-10}. This naming convention indicates that the model employs feedforward prediction with a query-centric representation, uses the Wasserstein distance as the distribution shift measure, and incorporates 10 modes. The naming scheme for other models used in the ablation studies follows a similar structure.

We consider several baseline approaches to identify interactive scenarios for comparison. First, we examine single-agent rules, including maximum velocity (\textbf{Rule-vel}) and maximum acceleration (\textbf{Rule-acc}). Additionally, we analyze agent-to-agent interaction rules, such as the minimum distance to surrounding agents (\textbf{Rule-dist}) and the number of agents within a 5-meter radius of the ego vehicle (\textbf{Rule-num}). Safety-related metrics, such as Time-to-Collision (TTC)~\cite{lee1976theory} and Time-to-Closest-Encounter (TTCE)~\cite{eggert2014predictive}, are also incorporated and referred to as \textbf{Rule-ttc} and \textbf{Rule-tcce}, respectively. Lastly, we consider two reconstruction-based metrics, commonly employed to measure the difficulty~\cite{makansi2021exposing} of a scenario for a planner: the discrepancy between the ground-truth future of the ego vehicle and a lane-following constant velocity planner (\textbf{Rule-lane}) and the deviation between the ground-truth future of all agents and the predictions generated by a learned model  (\textbf{Rule-err}).

\begin{figure}[t]
\centering
\includegraphics[width=0.99\linewidth]{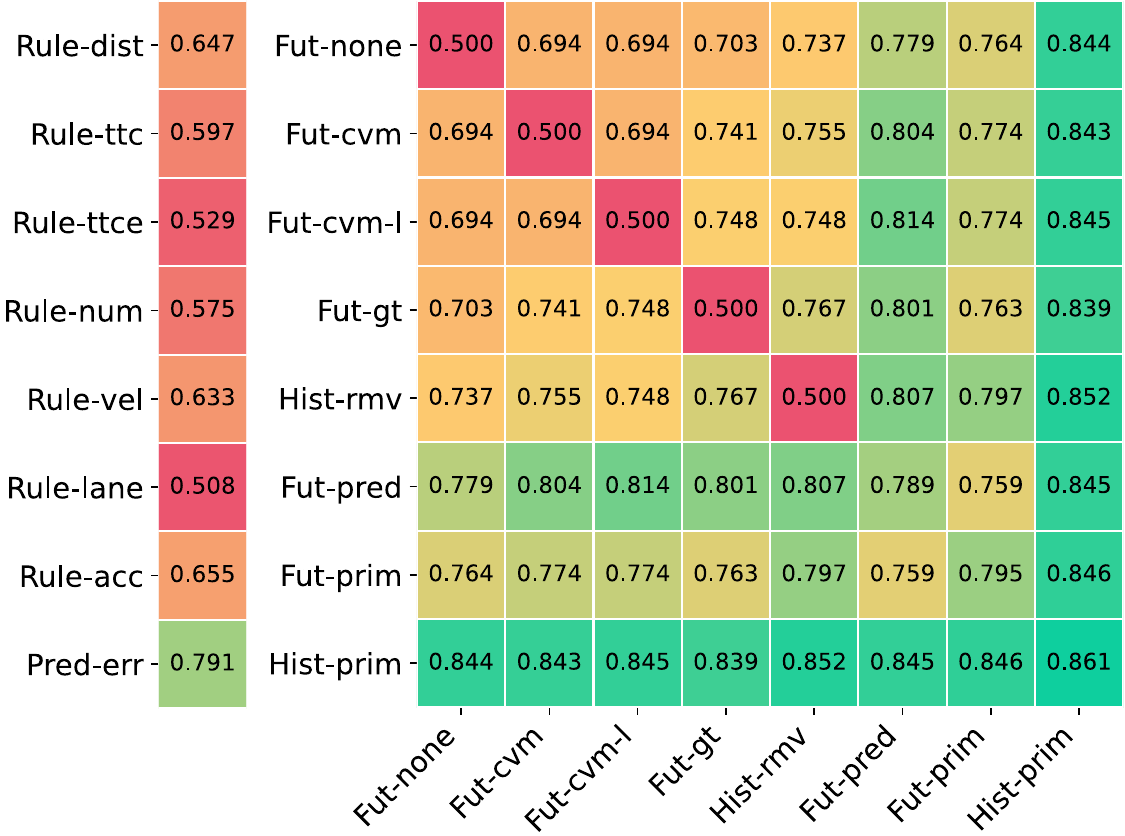} 
\vspace{-7mm}
\caption{The area under the Receiver Operating Characteristic curve (AUC-ROC) for various metrics used to assess interactivity, with higher values indicating better performance.}
\vspace{-5mm}
\label{fig:matrix_roc}
\end{figure}

\begin{figure*}[t]
\centering
\includegraphics[width=0.99\linewidth]{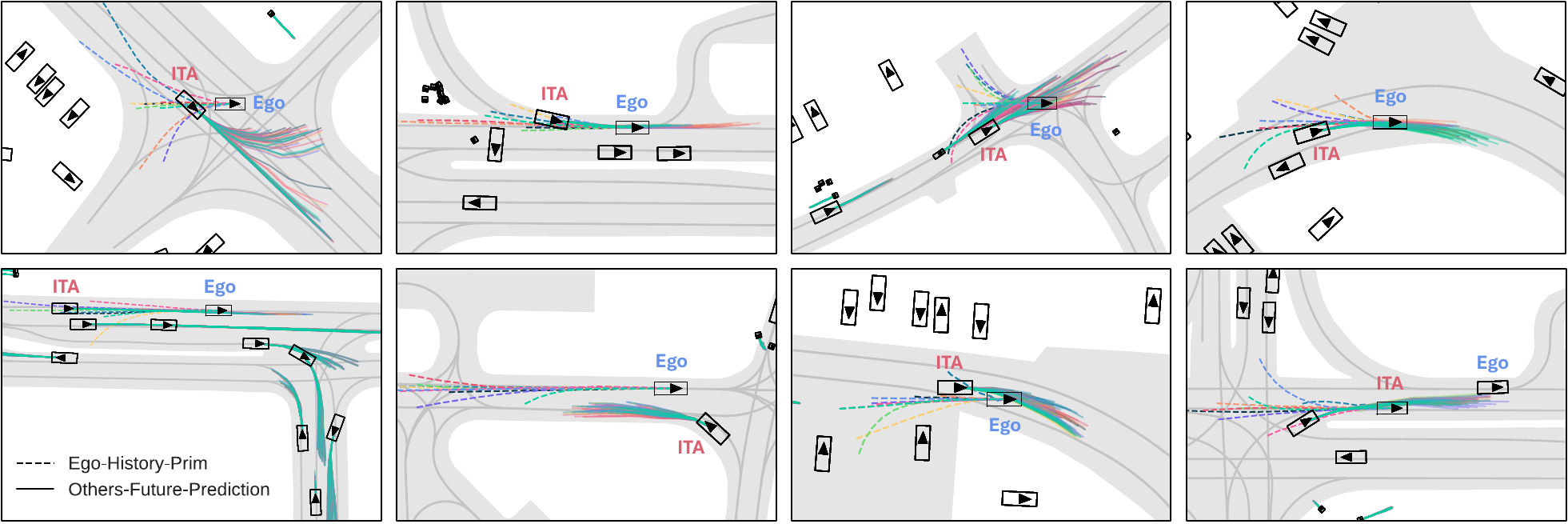}
\vspace{-1mm}
\caption{Examples of future predictions using \textbf{Hist-prim}, where different colors represent different counterfactual scenarios. Dashed lines indicate the history primitives of the ego vehicle (\textcolor{myblue}{Ego}), while solid lines represent the multi-modal predictions for other agents. With varying history primitives, interactive agents (\textcolor{myred}{ITA}) exhibit distinct future predictions, whereas non-interactive agents produce similar predictions.}
\vspace{-4mm}
\label{fig:example}
\end{figure*}

\begin{figure}[t]
\centering
\includegraphics[width=1.01\linewidth]{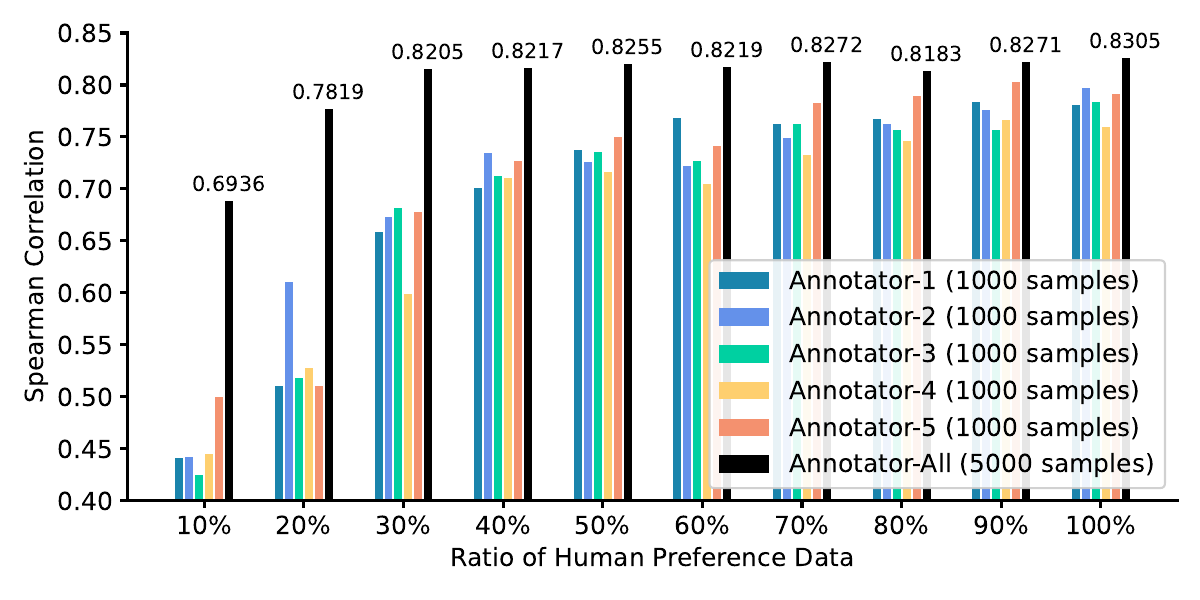}
\vspace{-8mm}
\caption{The impact of the human preference dataset size on training the reward model is illustrated by the correlation scores, which stabilize after utilizing $30\%$ of the dataset.}
\vspace{-6mm}
\label{fig:annotator}
\end{figure}

\subsection{Evaluation with Human Preference Label}

To assess the effectiveness of each counterfactual generation method, we examine its correlation with human preference annotations on the nuScenes dataset. We designed an online collection tool and asked each annotator to label 1,000 pairs following the instruction: \textit{select the more interactive scenario from two given scenarios}. The labels from five annotators were aggregated to train a reward model built on a motion prediction framework. Using this reward model, we ranked all samples in the validation dataset. Next, we computed the Spearman correlation~\cite{spearman1961proof} between the rankings predicted by the surprise metric and the ground-truth rankings generated by the reward model. Figure~\ref{fig:matrix_corr} presents the correlations for both the baselines and the surprise instances, with the x-axis representing the nominal scenario $\xi$ and the y-axis representing the counterfactual scenario $\mathcal{G}(\xi)$.

The results show that all surprise metrics exhibit strong positive correlations, demonstrating that the proposed surprise potential is an effective indicator of interactivity. In comparison, rule-based methods achieve correlations below 0.3. Notably, utilizing \textbf{Hist-prim} as the nominal distribution consistently yields correlations exceeding 0.8. This highlights \textbf{Hist-prim} as a strong surprise signal to capture interactivity in driving scenarios.

We further perform a binary classification analysis by categorizing the top $10\%$ ranked samples as interactive scenarios and the rest as non-interactive scenarios. The Area Under the Receiver Operating Characteristic (AUC-ROC) is shown in Figure~\ref{fig:matrix_roc}, a common metric for assessing model performance as the classification threshold is adjusted. The results indicate that \textbf{Hist-prim} remains the best-performing method, achieving an $86\%$ probability of correctly ranking a positive instance higher than a negative one~\cite{hanley1982meaning}.

\subsection{Analysis of Human Annotation}

To ensure the reasonableness of the ground-truth ranking labeled by the reward model, we investigate the impact of varying the dataset size used for training. In Figure~\ref{fig:annotator}, we train separate reward models for each annotator and a combined model using data from all annotators. The results show that the correlations rapidly saturate after using $70\%$ of the human preference dataset for training. Furthermore, the correlation stabilizes even when using only $30\%$ of the mixed data, suggesting that the dataset provides sufficient labels to effectively generalize human preferences for the nuScenes scenarios.

\subsection{Qualitative Evaluation}

To gain deeper insights into the best counterfactual generation method, \textbf{Hist-prim}, we present qualitative results in Figure~\ref{fig:example}. The history primitives are illustrated as dashed lines in various colors. These primitives serve as inputs to the prediction model, which generates the future trajectories of all other agents. The predicted trajectories are displayed as solid lines, matching the color of their respective history primitives.
The results reveal that interactive agents exhibit diverse future predictions, whereas non-interactive agents produce similar future predictions, even when the ego agent's history undergoes significant changes.

\begin{table*}[ht]
\caption{Spearman correlation score of ablation models.}
\label{tab:arch_dist}
\centering
\begin{tabular}{l|l|cccccccc|c}
    \toprule
    Ablation     & Method     & Fut-none & Fut-cvm & Fut-cvm-l & Fut-gt & Fut-pred & Fut-prim & Hist-rmv & Hist-prim & Average \\
    \midrule 
    \multirow{3}{*}{Architecture} 
                 & FFP-SC-W2-10  & 0.680    & 0.683   & 0.684     & 0.680  & 0.682    & 0.681    & 0.665    & 0.703   & 0.682 \\
                 & Diff-SC-W2-10 & 0.610    & 0.607   & 0.607     & 0.621  & 0.598    & 0.563    & 0.604    & 0.602   & 0.602 \\
                 & Diff-QC-W2-10 & 0.718    & 0.716   & 0.716     & 0.715  & 0.711    & 0.710    & 0.722    & 0.722   & 0.716 \\
    \midrule
    \multirow{2}{*}{Shift measurement} 
                 & FFP-QC-L2-10  & 0.688    & 0.702   & 0.704     & 0.698  & 0.697    & 0.700    & 0.678    & 0.709   & 0.697 \\
                 & FFP-QC-KLD-10 & 0.781    & 0.748   & 0.751     & 0.746  & 0.744    & 0.761    & 0.773    & 0.791   & 0.762 \\
    \midrule
    \multirow{4}{*}{Number of mode} 
                 & FFP-QC-W2-1  & 0.804    & 0.803   & 0.805     & 0.807  & 0.805    & 0.781    & 0.787    & 0.822    & 0.802 \\
                 & FFP-QC-W2-5  & 0.813    & 0.811   & 0.810     & 0.814  & 0.812    & 0.791    & 0.801    & \textbf{0.829}    & 0.810 \\
                 & FFP-QC-W2-10  & {0.812} & {0.809} & {0.814} & {0.809} & {0.810} & {0.804} & {0.808} & {0.827} & 0.812 \\
                 & FFP-QC-W2-15 & \textbf{0.814}    & \textbf{0.813}   & \textbf{0.816}     & \textbf{0.814}  & \textbf{0.814}    & \textbf{0.807}    & \textbf{0.811}    & {0.828}    & \textbf{0.815} \\
    \bottomrule
\end{tabular}
\vspace{-4mm}
\end{table*}

\subsection{Impact of the Design Choices}

We now examine the impact of the remaining two key dimensions of the surprise metric: the architecture of the prediction model and the measurement of distribution shift. Additionally, we analyze the effect of the prediction model's uncertainty by varying the number of modes in the prediction head.  Table~\ref{tab:arch_dist} presents the Spearman correlation results for the ablation models, using the counterfactual \textbf{Hist-prim} as the nominal distribution.

\textbf{Prediction model architecture}.
We find that query-centric is crucial for achieving high performance, as it more effectively captures the relationships between agents by using relative position embedding~\cite{zhou2023query}. We also observe that the diffusion model performs worse than \textit{FFP}, likely due to the sample-based distribution (50 samples in our experiment), which introduces additional noise. However, a notable advantage of the diffusion model is its inherent support for conditional generation using gradient guidance, which does not use future trajectory as input during the training stage.
%enabling a high correlation when using \textbf{Fut}. In contrast, conditional prediction models within \textit{FFP} do not fully capture the conditioning information, resulting in a lower correlation with \textbf{Fut}.

\textbf{Measurement of distribution shift}.
Our results show that \textbf{L2 norm} performs significantly worse than \textbf{KLD} and \textbf{W2}, suggesting that measuring shifts within the distribution space is essential for predicting interactivity. Additionally, we observe that \textbf{KLD} is slightly less effective than \textbf{W2}, highlighting the importance of accounting for modality alignment when evaluating distribution shifts.

\textbf{Prediction model uncertainty}.
The observations indicate that the uncertainty in multi-modal prediction has significant influences. To investigate this, we conducted experiments with varying numbers of modes in the prediction head. As shown in Table~\ref{tab:arch_dist}, the performance remains relatively consistent as the number of modes increases from 1 to 15. This suggests that even when the multi-modality property is not fully captured, the distribution shift continues to correlate with interactivity, demonstrating the robustness of our surprise metric.

\begin{figure}[t]
\centering
\includegraphics[width=0.99\linewidth]{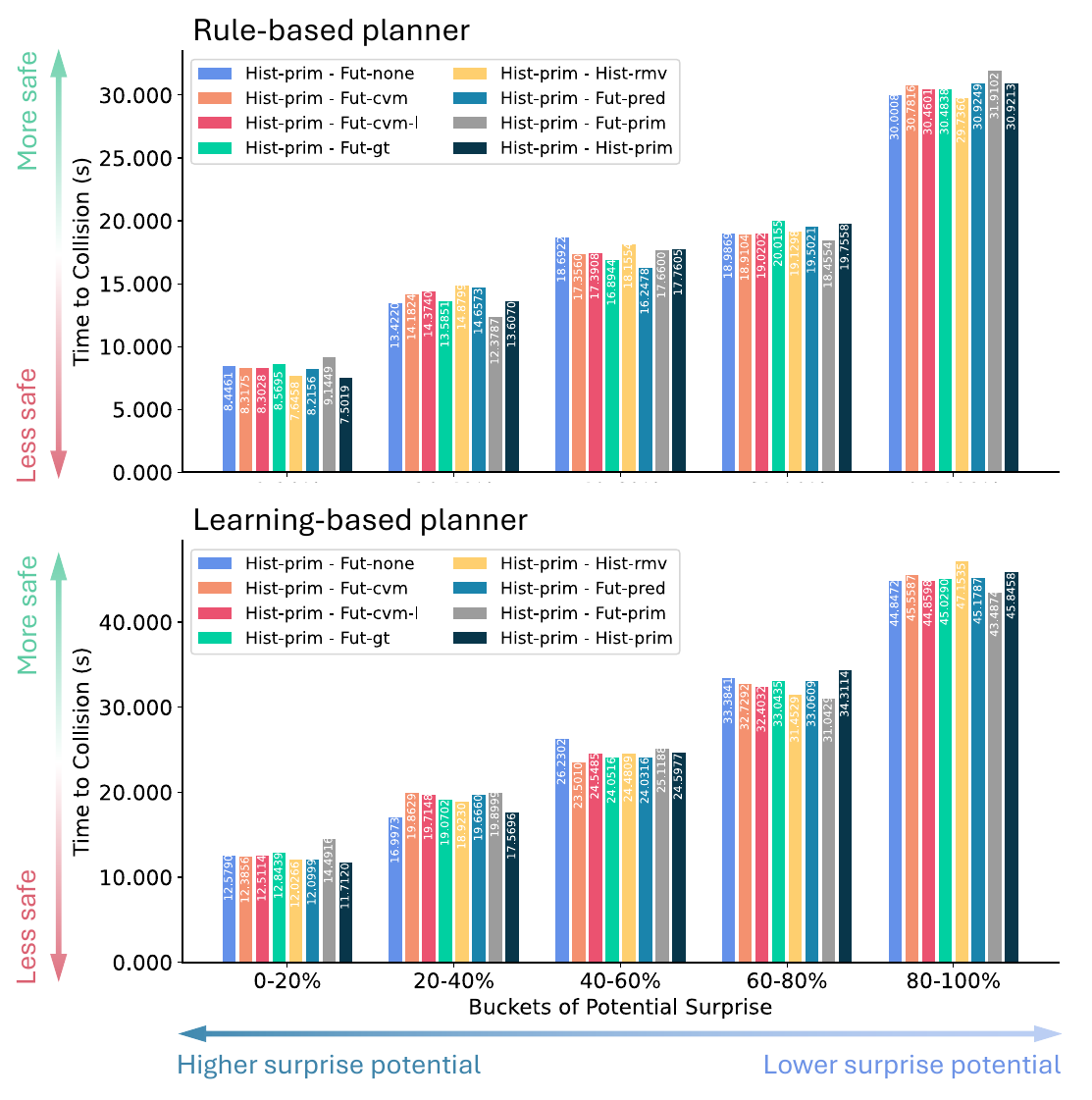}
\vspace{-7mm}
\caption{The time-to-collision results on different portions of the nuScenes validation dataset. The ranking is based on the surprise score generated by Hist-prim-Hist-prim.}
\vspace{-4mm}
\label{fig:planner}
\end{figure}

\section{Downstream application of curated interactive dataset}\label{sec:downstream}

We leverage our surprise potential metric to identify interactive scenarios and assess the utility of a curated interactive dataset in two downstream tasks: (i) planner evaluation and (ii) model training. Through these applications, we aim to further elucidate the importance of obtaining interactive driving data.

\subsection{Planner Evaluation on Interactive Dataset}

We evaluated two planners on the validation dataset and observed a clear relationship between the surprise potential and the planners' safety performance. The observed relationship indicates that the surprise potential score serves as a reliable proxy for identifying scenarios where a planner may underperform.

To investigate this, we divided the validation dataset into five equal buckets based on the surprise potential scores, corresponding to the top 0--20\%, 20--40\%, 40--60\%, 60--80\%, and 80--100\% of the data. We then evaluated two planners on each bucket: a rule-based planner, which maintains the current lane and velocity, and a learning-based planner, which generates the ego agent's plan using a motion prediction model. The evaluation focused on the mean Time-to-Collision (TTC)~\cite{westhofen2023criticality}, a widely used metric for assessing collision risk. The results, illustrated in Figure~\ref{fig:planner}, reveal a clear positive trend in surprise potential and TTC---the TTC decreases as the surprise score increases for both planners, indicating that high-surprise scenarios are more difficult and safety-critical.

\subsection{Model Training with Interactive Dataset}

\begin{figure}[t]
\centering
\includegraphics[width=0.99\linewidth]{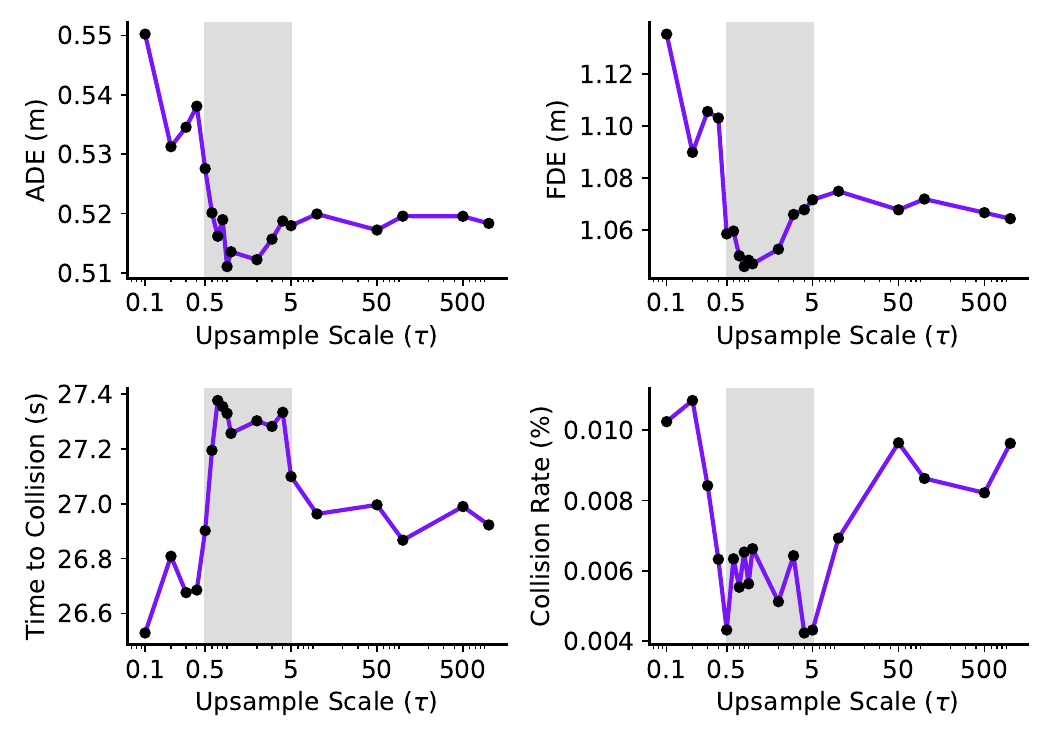}
\vspace{-8mm}
\caption{A planner was trained using different sampling weights, with performance evaluated using several metrics. For ADE, FDE, and collision rate, lower values indicate better outcomes, while for TTC, higher values are preferable. The gray region represents the range where the model achieves good performance for $\tau$ within [0.5, 5].}
\vspace{-4mm}
\label{fig:upsample}
\end{figure}

The discovered interactive scenarios can also be utilized for model training by upsampling the training dataset using weights for each sample defined as $w_i = \exp{[-r_i/(\tau N)]}$, where $r_i$ represents the ranking based on the surprise score, $N$ is the total number of samples, and $\tau$ controls the weight distribution. When $\tau \rightarrow \infty$, the sampling process reduces to a uniform distribution. Using different values of $\tau$, we trained multiple learning-based planners, evaluated them on the validation dataset, and presented the results in Figure~\ref{fig:upsample}.

Overall, the models demonstrate improved performance across all four metrics when $\tau$ is in the range of [0.5, 5]. When $\tau$ is less than 0.5, the model is trained on a dataset that lacks diversity and has an over-representation of a handful of challenging scenarios, thereby, it over-fits to these scenarios resulting in degraded performance.
% the model over-fits to a few challenging cases due to the dataset’s small size, leading to degraded performance. 
Conversely, when $\tau$ exceeds 5, the sample weights approach a uniform distribution, making the training dataset resemble the original dataset.

Although upsampling is effective, more advanced techniques, such as curriculum learning~\cite{bengio2009curriculum}, have the potential to further improve model training. 
In a data-driven world, the focus is gradually shifting from algorithm design to data workflows.
However, exploring these methods falls outside the scope of this paper and is left for future research.

\section{Key Takeaways}

Based on our experimental results, we summarize our findings via the following key takeaways.

\subsection{Does surprise potential indicate interactivity?}

Based on evaluating various surprise potential instantiations using human preferences, as shown in Figures~\ref{fig:matrix_corr} and \ref{fig:matrix_roc}, we find a strong correlation between the interaction in driving scenarios and the surprise. This suggests that the concept of surprise potential serves as a reliable indicator of interactivity.

\subsection{How to explain the gap between different counterfactuals?}

We observe that certain counterfactual choices yield better performance than others. Specifically, counterfactuals that perturb history generally outperform those that perturb futures. One possible explanation is that future-conditioned prediction models may suffer from information leakage, leading to overfitting on nominal and simple future scenarios. As a result, diverse and complex counterfactuals are likely treated as out-of-distribution during inference. Another factor contributing to the success of the \textbf{Hist-prim} instance is that its primitives encompass a wide range of behaviors, enabling the exploration of a larger and more diverse future space.

We also offer a semantic interpretation of \textbf{Hist-prim} from a causal perspective~\cite{pearl2009causality}.
A driving scenario can be represented as a structured causal model (SCM), where the history and future states of traffic participants are depicted as nodes, and the edges between them represent causal relationships. Perturbing the ego's history corresponds to an intervention operation, where different values are assigned to the ego history node, effectively removing the causal link influencing that node. Subsequently, the values of the future nodes are inferred based on the SCM, which is approximated by the motion prediction model.

\subsection{How does the model architecture influence?}

The results in Table~\ref{tab:arch_dist} indicate that query-centric representation plays a crucial role in performance, likely due to its ability to capture scene-level information effectively. In contrast, diffusion models, lacking an explicit probabilistic prediction head, are limited by the sample-based distribution. This limitation introduces additional noise into the shift measurement, resulting in a significant performance drop.

The ablation study on the number of prediction modes suggests that multi-modality is not a critical factor for measuring interactivity. Even with a single mode, the distribution shift between different counterfactuals was substantial enough to be noticeable.

\subsection{How does the shift measurement influence?}

We observe that \textbf{W2} and \textbf{KLD} outperform \textbf{L2 norm}, suggesting that accounting for uncertainty is crucial for identifying interactivity. The superior performance of \textbf{W2} compared to \textbf{KLD} can be attributed to the coupling process inherent in the Wasserstein distance for GMMs, which mitigates the noise introduced by mode permutations in different future scenarios.

\subsection{Are the discovered interactive scenarios useful?}

Based on the findings presented in Section~\ref{sec:downstream}, we conclude that the curated interactive scenarios are effective for both evaluating and training downstream models. 
Initially, we demonstrate that the surprise metric is a valuable tool to identify safety-critical scenarios. Furthermore, by applying a straightforward upsampling strategy, we show that the curated dataset has the potential to enhance the safety-related performance of a learning-based planner.

\section{Conclusion and Limitations}

In this paper, we exhaustively study the surprise potential metric to measure the interactivity of driving scenarios. With a deep investigation of instances from different implementations across three design dimensions, we have a better understanding of this metric and find that using counterfactual history motion primitives with Wasserstein distance dominates the others. This metric achieves more than 0.82 correlation scores with human annotation.

We observe that most false-positive cases are caused by the unstable prediction model, which could be improved by developing a more robust prediction model that is resistant to small perturbations. 
The experiments in this paper are based on a small dataset, which can be scaled to provide more detailed evaluation results. In addition, we conducted preliminary testing to highlight the downstream utility of a curated interactive dataset using our surprise potential measure. An important follow-up work would be to curate a larger and more diverse interactive dataset, and provide it as a benchmark for various downstream tasks such as planner validation, robustifying learned models, and generating novel interactive scenarios for the general robotics field.

% \section*{Acknowledgments}
% We would like to thank all human annotators for their efforts to label the preference data.

\bibliographystyle{IEEEtran}
\bibliography{references}

\end{document}